\documentclass[a4paper,runningheads]{llncs}
\usepackage{graphicx}
\usepackage{amsmath}
\usepackage[utf8]{inputenc}
\usepackage{xcolor}
\usepackage{subfig}
\usepackage{hyperref}

\begin{document}

\title{Generative Adversarial Networks for Operational Scenario Planning of Renewable Energy Farms: A Study on Wind and Photovoltaic}
\titlerunning{GANs for Scenarios in Renewables}

\author{Jens Schreiber, Maik Jessulat, Bernhard Sick}

\institute{University of Kassel\\ \texttt{\{j.schreiber, mjessulat, bsick\}@uni-kassel.de}}
\maketitle

\begin{abstract}
    For the integration of renewable energy sources, power grid operators need realistic information about the effects of energy production and consumption to assess grid stability. 
    Recently, research in scenario planning benefits from utilizing generative adversarial networks (GANs) as generative models for operational scenario planning.
    In these scenarios, operators examine temporal as well as spatial influences of different energy sources on the grid. 
    The analysis of how renewable energy resources affect the grid enables the operators to evaluate the stability and to identify potential weak points such as a limiting transformer. 
     However, due to their novelty, there are limited studies on how well GANs model the underlying power distribution. 
    This analysis is essential because, e.g., especially extreme situations with low or high power generation are required to evaluate grid stability. 
    We conduct a comparative study of the Wasserstein distance, binary-cross-entropy loss, and a Gaussian copula as the baseline applied on two wind and two solar datasets with limited data compared to previous studies. 
    Both GANs achieve good results considering the limited amount of data, but the Wasserstein GAN is superior in modeling temporal and spatial relations, and the power distribution.
    Besides evaluating the generated power distribution over all farms, it is essential to assess terrain specific distributions for wind scenarios. 
    These terrain specific power distributions affect the grid by their differences in their generating power magnitude.
    Therefore, in a second study, we show that even when simultaneously learning distributions from wind parks with terrain specific patterns, GANs are capable of modeling these individualities also when faced with limited data. 
    These results motivate a further usage of GANs as generative models in scenario planning as well as other areas of renewable energy.
\end{abstract}

\section{Introduction}
\begin{figure}[htbp]
    \centering
    \includegraphics[width=0.98\textwidth]{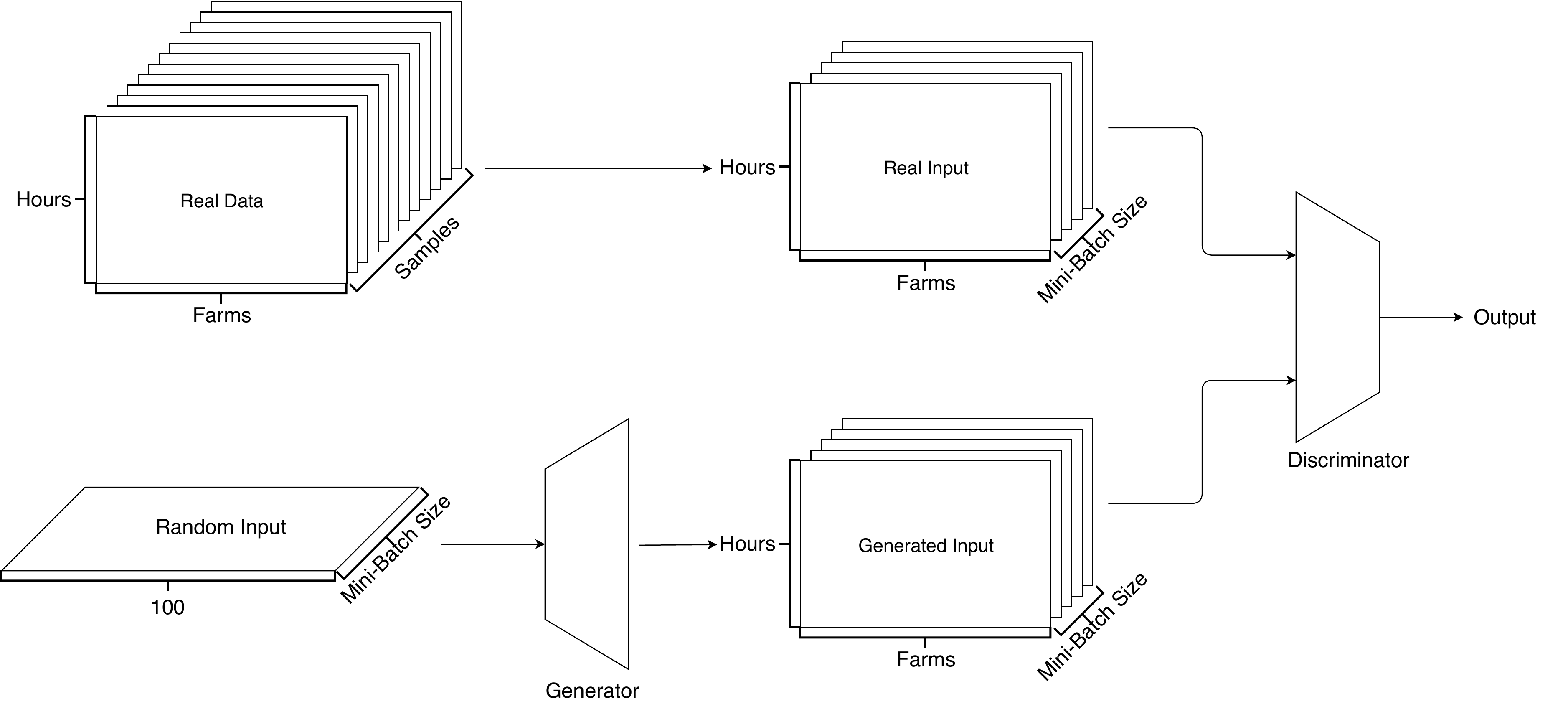}
    \caption{To train the GANs, samples are selected from the entire dataset according to the batch size. Then, the generator creates a second dataset of the same size from a random input. The shape of the datasets reflects the temporal and spatial relationship. Both datasets need to be distinguished by the discriminator. The output of the discriminator corresponds to how likely a sample of the batch is either real or generated.}
\label{input}
\end{figure}

Renewable energy sources are by now an essential energy producer of the electrical power grid~\cite{fraunhofer2018windenergie,Lowery2013}. 
By integrating these power plants, we introduce a lot of volatile energy. 
To maintain a stable power grid, power grid operators need realistic information about the effects of energy production and consumption assessing grid stability~\cite{Lowery2013,Jens2018}. 
It is essential to use operational scenario planning~\cite{Sovan2008,Lowery2013} to evaluate the integration of renewables.

Traditionally, generative approaches such as stochastic programming, copula methods, or Monte-Carlo approaches allow simulating the stochastic and intermittent nature of renewable power generation~\cite{Hart2011,Becker2018,Kaut2003Evaluation}. Often, these techniques only allow for modeling either the temporal or the spatial relationship of renewable energy sources.
More recently, research in scenario planning shows the strong capabilities of generative adversarial networks (GANs) modeling the temporal as well as the spatial relationship of, e.g., wind and photovoltaic (PV) farms~\cite{Chen2018Bay,Chen2018ModelFree,Chen2018Unsup} simultaneously. 

Utilizing GANs for scenario planning is especially interesting because it allows for simulating a large number of realistic power samples after initial training of the GAN. Further, the training, see Figure \ref{input}, emphasizes the spatial relation (of different farm locations) as well as temporal relation (of the simulated hours) through using historical data.
However, due to its novel application in the field, there are limited studies on how well GANs can model the underlying power distribution, especially with limited data for training. This analysis is essential because, e.g., extreme situations with low or high power generation are required to evaluate the grid stability.

Besides evaluating the generated power distribution over all farms, it is essential to assess terrain specific distributions, caused by location-specific weather conditions, for wind power scenarios. Often, the average power generation and also the density associated with power values are different for different terrains~\cite{Pinson2006,jens2019}. To evaluate these effects on the grid in various terrains, the analysis on GANs to generate those distributions is essential.

\noindent
Therefore, the main contributions can be summarized as follows\footnote{Implementation details of the evaluation, the experiment, and the training is available at \url{https://git.ies.uni-kassel.de/scenario_gan/scenario_gan_wind_pv}.}:

\begin{itemize}
    \item We provide a comparative study of two different loss functions (binary-cross-entropy loss and Wasserstein distance) on two solar and two wind dataset (with limited historical data for training) to evaluate the underlying power distributions through the Kullback-Leibler divergence (KLD).
    \item Results show that the Wasserstein distance is superior over the binary-cross-entropy and a Gaussian copula (GC) baseline even when faced with limited data compared to previous studies.
    \item A study on how location-specific influences and weather conditions (that affect the power distribution) shows that GANs learn those specifics even when only four offshore parks are present in the dataset.
\end{itemize}

The remainder of this article is structured as follows. In Section~\ref{relatedwork}, we give an overview of related work. Section~\ref{methodology} describes two types of loss functions and details the evaluations measures. We continue by describing the experiments and results in Section~\ref{experiment}. Section~\ref{conclusion} summarizes the article and provides an outlook on future work.

\section{Related Work}
\label{relatedwork}
The evaluation of grid stability through operational scenario planning is an essential research topic to integrate volatile renewable energy resources.
By creating realistic realizations of stochastic processes in the field of renewable energy, we can analyze their potential impact in real-world scenarios~\cite{Antonio2010}.
Often, generating scenarios in the field of renewable energy is done by techniques provided by stochastic programming, copula methods, or Monte-Carlo approaches. In the following, we give a brief overview of these techniques, followed by a summary on utilizing GANs for simulating scenarios.

Already in 2013,~\cite{Kaut2003Evaluation} developed evaluation methods and algorithms for using stochastic programming in scenario simulations. Further, the authors present a stochastic programming method for portfolio management.
Various probabilistic prediction methods are also used to simulate scenarios. These have the advantage that they already model the distribution of power. Besides, they allow modeling temporal relationships as given by the prediction models. This modeling enables the authors in~\cite{Pinson2009} to provide a method for converting probabilistic predictions into multivariate Gaussian random variables. 
In particular, it focuses on the simulation of scenarios which model the interdependent temporal effects from prediction errors. 
It is also possible to create scenarios based on probabilistic predictions~\cite{Iversen2016}. 
For the evaluation of such scenarios,~\cite{Pinson2012} defines criteria measures, such as the energy score, and gives recommendations. However, the implementation of the score is tedious and error-prone.

The authors in \cite{Becker2018} use a copula approach to model temporal affects onto forecasts with distinct forecast horizons allowing to distinguish between the uncertainty in wind power forecasts and temporal dependencies. 
The presented method outperforms all other approaches in their experiments.
Recent work in ~\cite{TaoWang2016} presents a new proposal that models spatial dependence between renewable energy resources. Therefore, the implemented prototype uses Latin hypercube sampling and copula methods and is tested on actual wind power measurements and power forecasts.

A comprehensive study on real-world data in \cite{Rachunok2018} shows the trade-off between computational complexity and the quality of simulated scenarios when using Monte-Carlo techniques. Another method uses a Monte-Carlo approach~\cite{Hart2011}, to study a planning tool that takes various renewable resources from different locations into account — further, the authors consider temporal effects in simulations for load scenarios.

Most of the previous literature either considers temporal or spatial effects. However, recently, utilizing GANs allows simulating wind and solar scenarios that take spatial and temporal relations into account~\cite{Chen2018ModelFree}. 
They also show how to create scenarios with wind ramp events by utilizing conditional GANs. 
It is shown in \cite{Chen2018Unsup}, that GANs are capable of simulating scenarios conditioned on a previous forecast.
In \cite{Chen2018Bay}, Bayesian GANs create realistic scenarios for wind and PV simultaneously. In a sense, the approach in \cite{Chen2018Bay} is similar to ours, as we show the capability of GANs to simulate parks of different terrains together.

The literature review shows that most of the work is focusing on either the temporal or spatial evaluation. Further, a comparison between the historical data and the generated data distribution is not provided using known measures such as the KLD. Besides, none of the articles presents an analysis of whether it is possible to create terrain specific power distribution when simultaneously simulating power distributions of numerous wind farms.
Further, previous studies have a large amount of data, e.g., \cite{Chen2018ModelFree} uses $14.728.320$ measurements, compared to our datasets with a maximum of $490.752$ historical power measurements as detailed in Section~\ref{sec:data}.

\section{Methodology}
\label{methodology}
After giving a short introduction into the applied GANs, we detail methods to evaluate the simulated power distribution with the distribution from historical data.

\subsection{Generative Adversarial Networks}
GANs consist of two different neural networks~\cite{goodfellow2014generative}: The discriminator and the generator. 
In Figure~\ref{input}, the generator takes some random values and produces fake samples to imitate the distribution of a real dataset. 
This imitation enables us to make use of spatial and temporal relations already present in historical data.
The discriminator, on the other hand, takes real and fake samples as its input and tries to distinguish between real and generated samples.
During the training, the quality of the generated data, as well as the classification accuracy of the discriminator, should increase.
The improvement depends on the loss functions used.
After training, the generator produces examples from the distribution of the original data. 
The discriminator, whereas, can detect novelties and outliers in the data~\cite{Zenati2018}.
Often, GANs employ the Wasserstein distance~\cite{arjovsky2017wasserstein} or the binary cross entropy (BCE)~\cite{radford2015unsupervised} as loss function. Later on, we refer to the network with the BCE loss function as deep convolutional GAN (DC-GAN) and deep convolutional Wasserstein GAN (DC-WGAN) as the network trained with the Wasserstein distance.

The BCE is defined as follows:
\begin{equation*}
BCE = -\left(y \log \left(p\right) + \left(1-y\right) \log \left(1-p\right) \right), \\
\end{equation*}
where $y$ stands for the label if the data is real or generated, and $p$ for the probability that the discriminator assigns (given by the sigmoid function at the final layer). A zero label means that the data is classified as generated, while a one corresponds to the real data. 

The Wasserstein distance \cite{arjovsky2017wasserstein} is a measure that is used to compare two distributions. 
It is also referred to as earthmover distance and indicates the effort that is required to transform one probability distribution into another distribution.
It is defined as follows
\begin{equation}
W_p\left(\mu, \nu\right) = \inf E\left[d\left(X, Y\right)^p\right], 
\label{wasserstein_metric}
\end{equation}
where $X$ and $Y$ are the distributions in the range between $\mu$ and $\nu$.
Since there is not only one possible solution to convert one distribution into another, the solution chosen for this loss is the one with the least effort, which corresponds to the infimum (inf) in Equation~\ref{wasserstein_metric}.

\subsection{Kernel Density Estimation}
The kernel density estimation (KDE) is a statistical method to determine the distribution of a given dataset. In the KDE algorithm, superimposing several Gaussian distributions allows for estimating the probability density function (PDF) for datasets. Applying KDE to the historically measured and generated power data allows comparing them with each other, e.g., by employing the Kullback-Leibler divergence.

\subsection{Kullback-Leibler Divergence}
The KLD is a non-symmetric statistical measure to determine the difference between the distributions. Later on, we use the KLD to quantify the similarity between the generated and historical data through a KDE.
It is defined as
\begin{equation}
D_{KL}\left(P||Q\right) = \int_{-\infty}^\infty p\left(x\right) \log \left(\frac{p\left(x\right)}{q\left(x\right)}\right) dx
\end{equation}
with $P$, $Q$ as the distributions and $p(x)$,  $q(x)$ as their PDFs. Due to the non-symmetrical behavior, both $D(P||Q)$ and $D(Q||P)$ are calculated and added together.
One interpretation of the  KLD is as information gain achieved by replacing distribution $Q$ with $P$.

\section{Experimental Set-Up and Evaluation}
\label{experiment}
This section presents the experimental set-up and evaluation results. 
Therefore, we detail the different datasets and explain the preprocessing of the data.  
Further, we describe the architectural set-up of the evaluated DC-GAN and DC-WGAN. 
Afterward, we evaluate theses GANs concerning a GC baseline.
In particular, we evaluate the generated samples regarding their temporal and spatial correlation, their generated distribution, and the creation of high and low-stress power profiles. 
In the final study on the GermanWindFarm2017 dataset, we assess how different terrains and their location-specific wind conditions (that affect the power distribution) are modeled by the DC-WGAN when trained simultaneously.

\subsection{Data}\label{sec:data}

The \textit{EuropeWindFarm2015} and \textit{GermanSolarFarm2015} dataset can be obtained via our website\footnote{\url{https://www.ies.uni-kassel.de}}. 
We further use a \textit{GermanWindFarm2017} and \textit{GermanSolarFarm2017} dataset, which are not publicly available. 
However, especially the GermanWindFarm2017 dataset allows us to get additional insights into the power distribution relating to terrain-specific conditions. 
These datasets make our data quite diverse and we cover a broad spectrum of power distribution from the wind as well as solar problems. 

Compared to previous studies on GANs for renewable power generation, see, e.g., \cite{Chen2018Unsup,Chen2018ModelFree} with a total of $14.728.320$ measurements and a five-minute resolution, we only have a limited amount of data. 
The largest of our datasets has $490.752$ power measurements, as detailed in Table~\ref{tbl:datasets}. 
The solar datasets have a three-hourly resolution totaling in $8$ measurements per day. 
Wind datasets have an hourly resolution with $24$ power measurements per day.

To discover relations within the data, we aim at making spatial and temporal relationship available in each training sample. 
Therefore, we reshape the data to obtain a $P\times H$ shaped matrix for each day (sample), where $P$ refers to the number of parks and $H$ refers to the time steps within the horizon. 
This matrix is obtained by first creating a list of samples for each farm with its respective time steps (horizon) and afterward combine all individual time steps of all farms.
Finally, the reshaping allows the utilized convolutional layers, see Section~\ref{sec:gan_training}, to make use of their \textit{receptive field} and discover relations within the data, either temporal or spatial. Respectively, the number samples in Table~\ref{tbl:datasets} refer to the number of matrices with shape $P\times H$.
\begin{table}
    \centering
    \begin{tabular}{|l||r|r|r|r|r|}
    \hline
    Name                     & \#Parks           & Resolution        &Horizon  & \#Samples  &\#Measuresments  \\
    \hline
    \hline
    EuropeWindFarm2015         & $32$              &$1$h            &$24$ time steps     & 540 & $414.720$\\
    GermanSolarFarm2015        & $16$              &$3$h          &$8$ time steps      & 760 &  $972.80$ \\
    GermanWindFarm2017         & $48$              &$1$h          &$24$ time steps     & 426 &  $490.752$\\
    GermanSolarFarm2017        & $48$              &$3$h          &$8$ time  steps      & 483 &  $185.472$\\
    \hline
    \end{tabular}
    \vspace{1em}
    \caption{Summary of the evaluated datasets. The samples refer to the number of matrices with shape $P\times H$ (the number of parks times the number of time steps within a datasets horizon).}
\label{tbl:datasets}
\end{table}





After normalizing and reshaping the data, we randomly select $80\%$ of the data for training and the remaining historical data for testing. 

\subsection{GAN Training}\label{sec:gan_training}

To discover relations within the data, the applied GANs are designed to make use of the receptive field of convolutional networks. Therefore, the generator utilizes convolutional layers to create samples of the form $P\times H$ subsequently. Depending on the dataset, the generators parameter (kernel size, stride, and padding) are selected to fulfill this requirement as detailed in Table~\ref{gan_parameters}. Varying stride and padding allow to almost consistently apply a kernel size of $4$ while achieving a receptive field sufficient to cover the complete $P\times H$ matrix.





\begin{table}
    \centering
    \begin{tabular}{|l||r|r|r|r|}
    \hline
    Dataset name                     & Kernel size         & Stride        & Padding     \\
    \hline
    \hline
    EuropeWindFarm2015       &  $[(4, 3), 4, 4, 4]$ & $[1, 2, 2, 2]$ & $[0, 1, 1, 1]$      \\
    GermanSolarFarm2015      &  $[(2, 1), 4, 4, 4]$ & $[1, 2, 2, 2]$ & $[0, 1, 1, 1]$      \\
    GermanWindFarm2017       &  $[3, 4, 4, 4]$ & $[1, 2, 2, (4, 2)]$ & $[0, 1, 1, (0, 1)]$      \\
    GermanSolarFarm2017      &  $[(3, 1), 4, 4, 4]$ & $[1, 2, 2, (4, 2)]$ & $[0, 1, 1, (0, 1)]$      \\ 
    \hline
    \end{tabular}
    \vspace{1em}
    \caption{Summary of the generator's configuration for each dataset: The discriminator's parameters are in reverse order. The output shapes of the generator are $[100, 256, 128, 64, 1]$ and in reverse order for the discriminator.}
\label{gan_parameters}
\end{table}

The discriminator's parameters are reverse to the generators. This (reverse) parameter set-up allows making best use of the joint training and the receptive field of the convolutional layers because the discriminator is capable of detecting missing relations in the generated data and on the other hand the generator is capable of creating those.

In the following, we evaluate two GANs trained with the Wasserstein distance (DC-WGAN) and the BCE loss (DC-GAN). We apply batch normalization inside the discriminator and the generator. As activation function, we use leaky Rectified Linear Units (ReLU). The GANs are trained for $50000$ epochs with a learning rate of $2\mathrm{e}{-5}$ and a batch size of $64$.

\label{evaluation}
\subsection{Study on EuropeWindFarm and GermanSolarFarm Dataset}
In this section, we highlight the results of the comparative study of the DC-GAN, DC-WGAN, and a GC~\cite{Nelsen2006} as the baseline.

\subsubsection{Evaluation through KLD:}

To evaluate the generated distributions, we apply a KDE to the test dataset and the samples created by the models. 
The KDE uses a Gaussian kernel, a Euclidean distance~\cite{scikit-learn}, and a bandwidth of $0.01$ to reflect $1\%$ of the normalized power. 
The PDFs from the KDE algorithm are used to examine the similarity between the distributions of real and generated data using the KLD. Table~\ref{Distributions2} summarizes this comparison of generated samples and historical power data. 
\begin{table}
    \centering
    \begin{tabular}{|l||c|c|c|c|}
    \hline
    Location                 & KLD GC           &KLD DC-GAN        &KLD DC-WGAN  \\
    \hline
    \hline
    EuropeWindFarm2015          & $0.068$              &$0.663$            &$ 0.029$   \\
    GermanSolarFarm2015         & $0.042$              &$0.011$            &$0.011$  \\
    GermanWindFarm2017          & $0.062$              &$0.218$            &$0.027$  \\
    GermanSolarFarm2017         & $0.034$              &$0.942$            &$0.008$  \\
    \hline
    \end{tabular}
    \vspace{1em}
    \caption{The table highlights the evaluation results, of the generated distributions after training the different GANs. The KLD is calculated between the created and real distributions for all farms together from each dataset for the DC-GAN, DC-WGAN and the GC.}
\label{Distributions2}
\end{table}
Results for all datasets show that the DC-WGAN is superior over the GC baseline. 
DC-GAN has worse results than GC for all datasets except the GermanSolarFarm2015.
The DC-WGAN creates samples with smaller, or at least a similar low, KLDs compared to the DC-GAN and the GC, showing its excellent performance.
Note that Figure~\ref{locations} provides a representative example of those distributions for the GermanWindFarm2017 dataset.

Interestingly, even though the limiting amount of training data compared to other studies, results of the KLD suggest that generated samples reflect the distribution of the real world. These positive results are potentially due to the combined training and the selected parameters to make the best use of the receptive field.

\subsubsection{Evaluation of Temporal and Spatial Relation:}
\begin{figure}[!tb]%
    \subfloat[Temporal relation in historical data.]{{\includegraphics[width=0.48\textwidth]{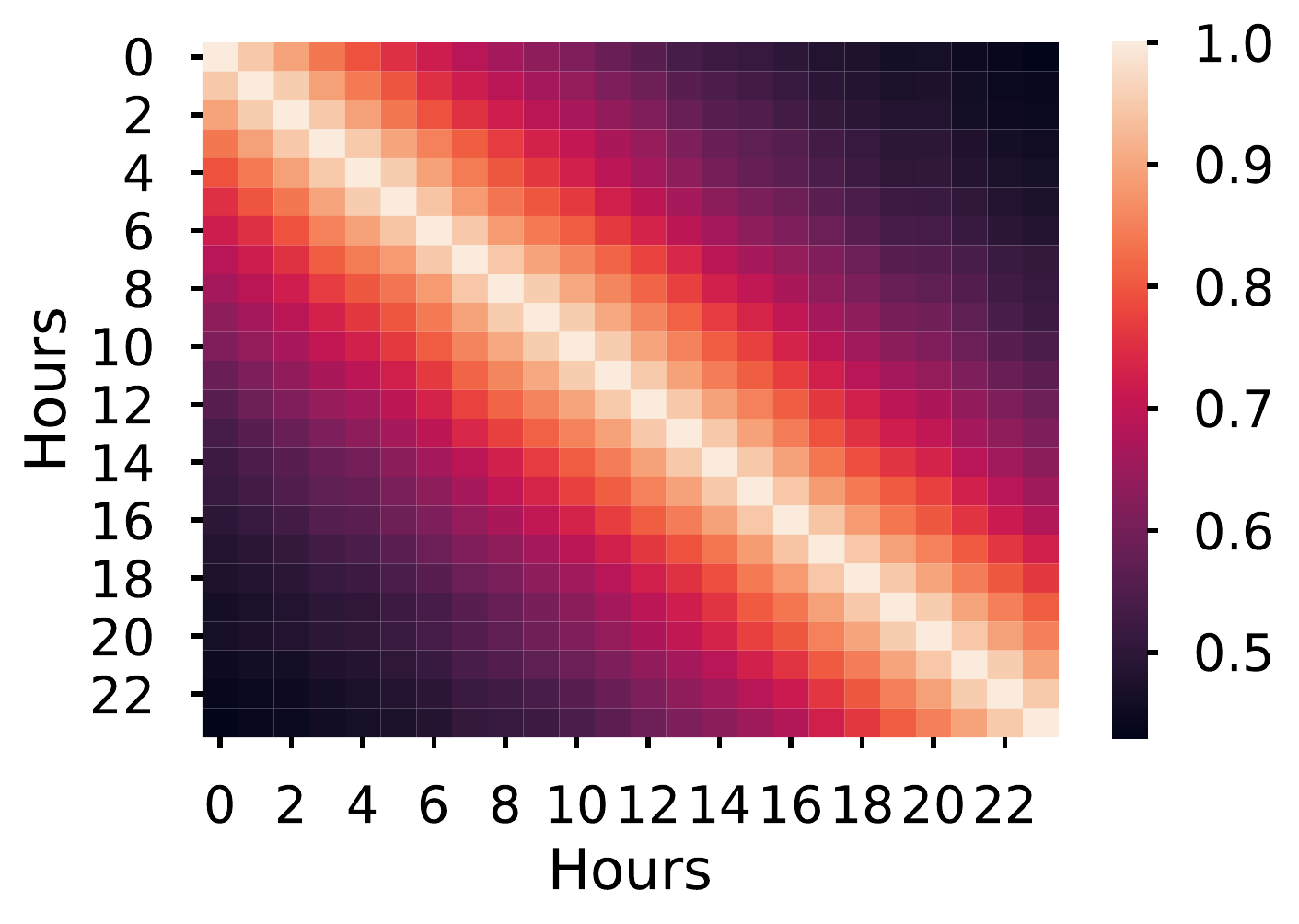} }}%
    \hfill
    \subfloat[Temporal relation in generated data from DC-WGAN.]{{\includegraphics[width=0.48\textwidth]{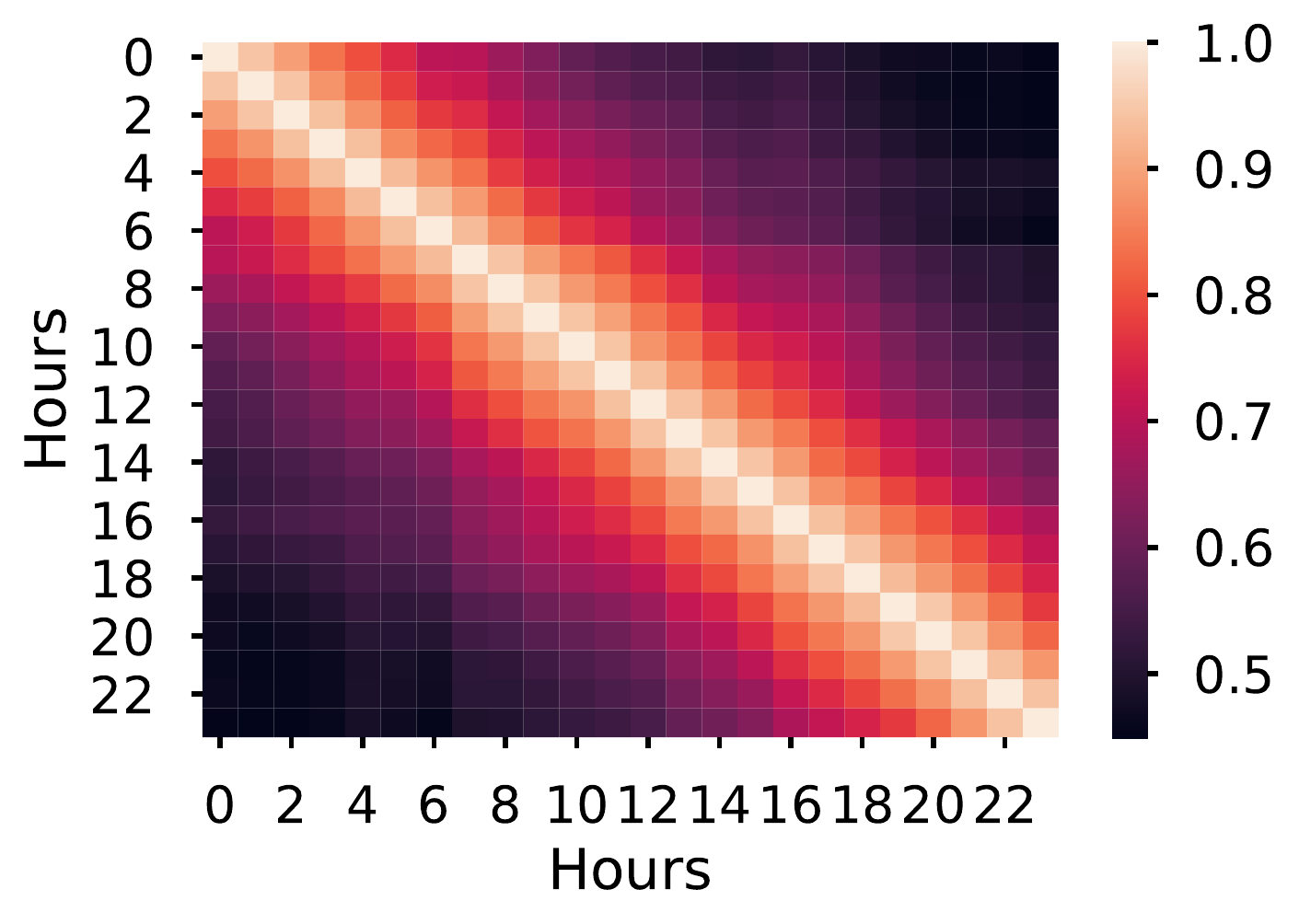} }}%
    \caption{The figures provide a comparison of the temporal relation present in the real-world data and the generated samples using Pearsons correlation matrix~\cite{Chen2018Unsup} of the EuropeWindFarm2015 dataset.}%
    \label{temporal}%
\end{figure}

Besides creating data of similar distribution, it is essential to examine the spatial and temporal relationship at the same time.
As results of the KLD suggest the superior performance of the DC-WGAN over the DC-GAN and GC baseline, we limit the following discussion to the DC-WGAN.
Nonetheless, note that results of the DC-GAN and GC are reasonable but outperformed by the DC-WGAN.
To restrict the discussion to relevant and non-repetitive results, we limit the following analysis to two representative examples that provide details due to their increased data availability. 
For example, the wind datasets provide more details about the temporal relation as these include more time steps compared to the solar datasets.

Figure~\ref{temporal} shows typical results using Pearson's correlation matrix to calculate temporal relations for the generated hours~\cite{Chen2018Unsup}. 
In these results for the EuropeWindFarm2015 dataset, we observe that in real-world as well as in the generated samples from the DC-WGAN, the power values have a higher Pearson coefficient for hours related to each other.

\begin{figure}[!tb]%
    \centering
    \subfloat[Spatial relation in historical data.]{{\includegraphics[width=0.48\textwidth]{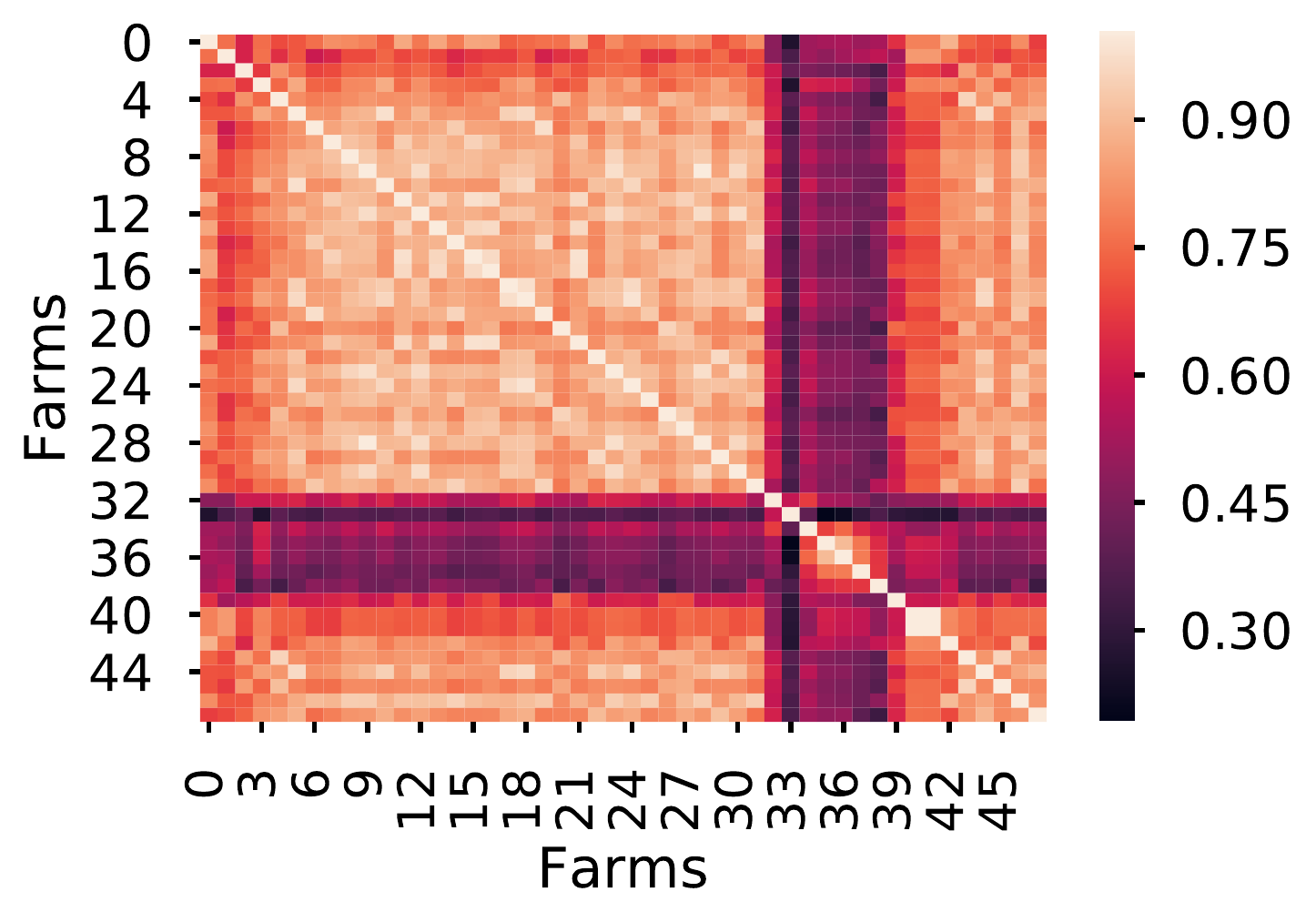} }}%
    \hfill
    \subfloat[Spatial relation in generated data from DC-WGAN.]{{\includegraphics[width=0.48\textwidth]{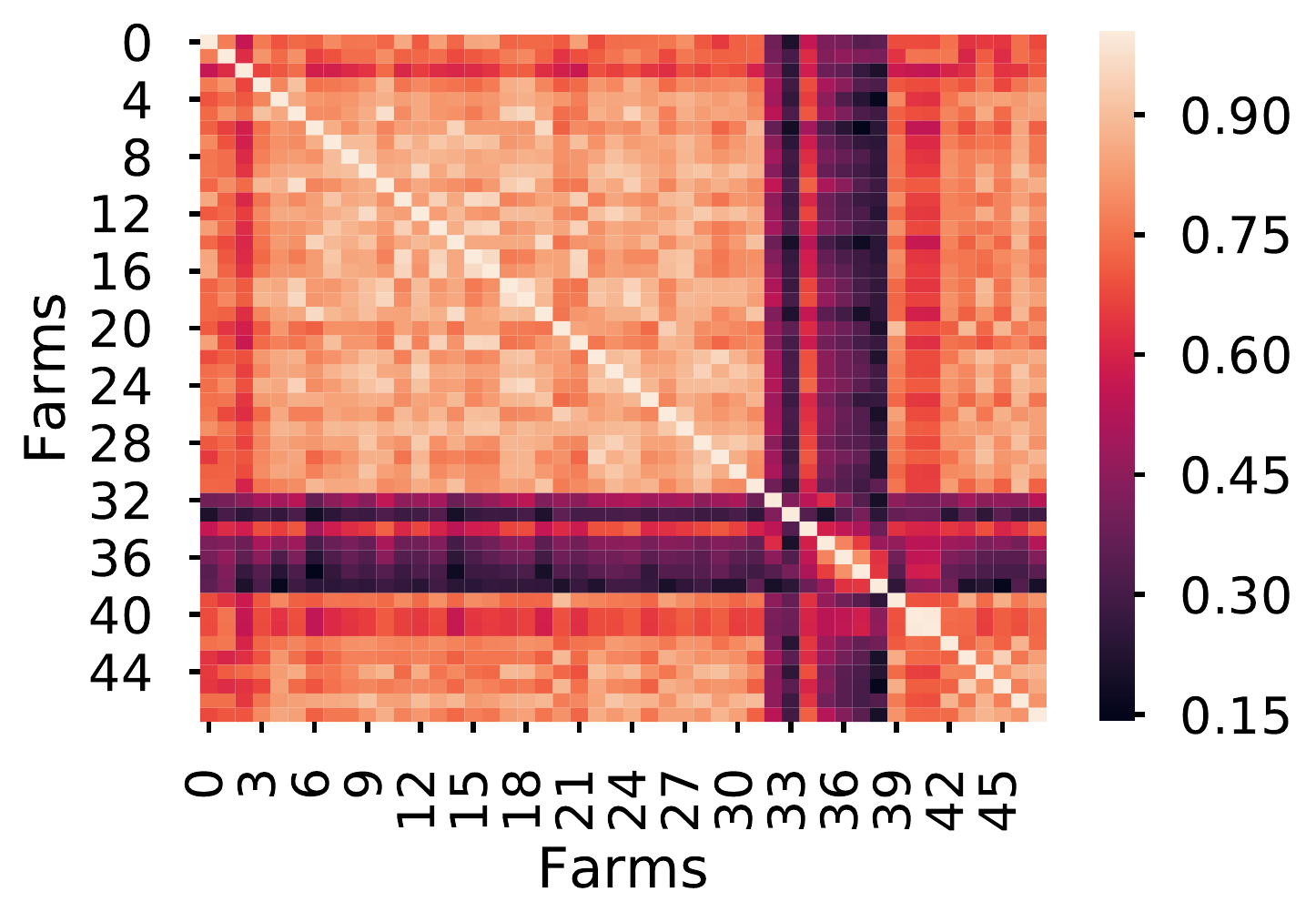} }}%
    \hfill
    \caption{Comparison of the spatial relation present in the real-world data and the generated samples using Pearsons correlation matrix~\cite{Chen2018Bay,Chen2018ModelFree} of the GermanSolarFarm2017 dataset.}%
    \label{spatial}%
\end{figure}

Figure~\ref{spatial} shows exemplary results using Pearsons correlation matrix to measure spatial relations between different farms~\cite{Chen2018Bay,Chen2018ModelFree}. 
In these results of the GermanSolarFarm2017 dataset, we observe that similar spatial relations, measured by the Pearson coefficient, are present on the historical data as well as the generated samples. 
In most cases, a high correlation is present and the DC-WGAN captures almost all those spatial relationship compared to the heatmap from historical data. 
In some examples, the DC-WGAN creates samples with a more substantial spatial relation to each other than present in the historical data.

Interestingly, for some farms, there is a rather modest spatial relation in the historical as well as the generated data. 
This relation is unlikely in the case of small regions (such as Germany) and can be caused by maintenance problems, shadowing effects, or other problems in the data.
As the presented results are representative for all datasets, the above results show that the DC-WGAN is capable of reconstructing the historical power distribution, the spatial relation, and temporal relation for all datasets even for a varying amount of farms, resolutions, and the number of historical power measurements for training.

\subsubsection{Evaluation of Generated Power Profile:}

To asses grid stability, the creation of different stress situations is essential. 
A typical stress scenario involves large power generation over a long period, as this causes the maximum thermal load on the elements and is therefore relevant for selecting the correct technical characteristics of those elements.

The following section gives insights into the amount of \textit{stress} by calculating the integral over the generated time horizons. 
E.g., for the wind datasets for each wind farm $24$ power values are generated corresponding to $24$ hours. 
As the maximum power normalizes the data, the maximum value of the integral is $24$ for a single farm of a sample.
The maximum value for the solar datasets is about $4$ because power is not created at night.

\begin{figure}[!tb]%
    
    \centering
    \subfloat[Integrated power generation from historical data.]{{\includegraphics[width=0.48\textwidth]{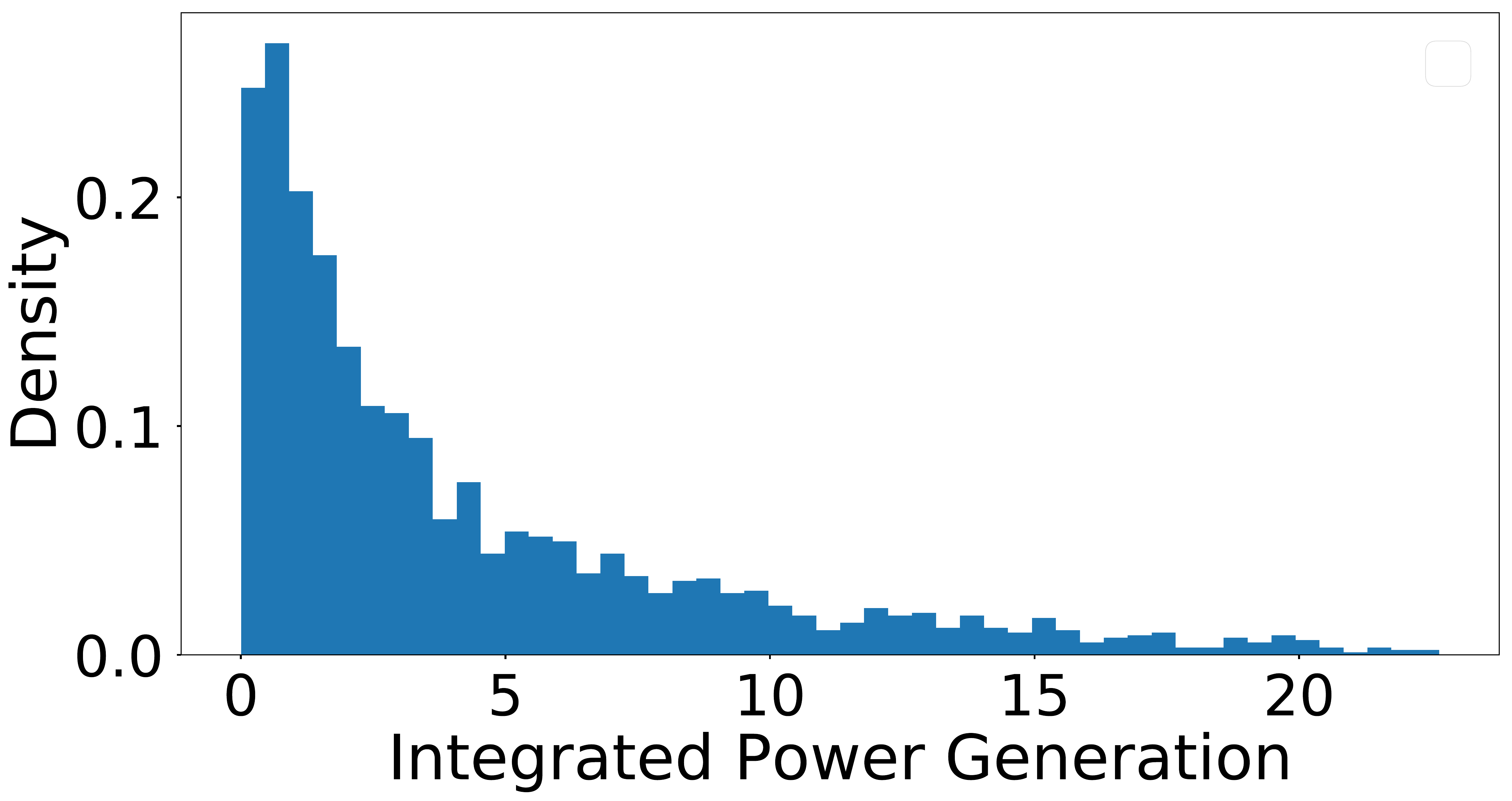} }}%
    \hfill
    \subfloat[Integrated power generation from generated DC-WGAN samples.]{{\includegraphics[width=0.48\textwidth]{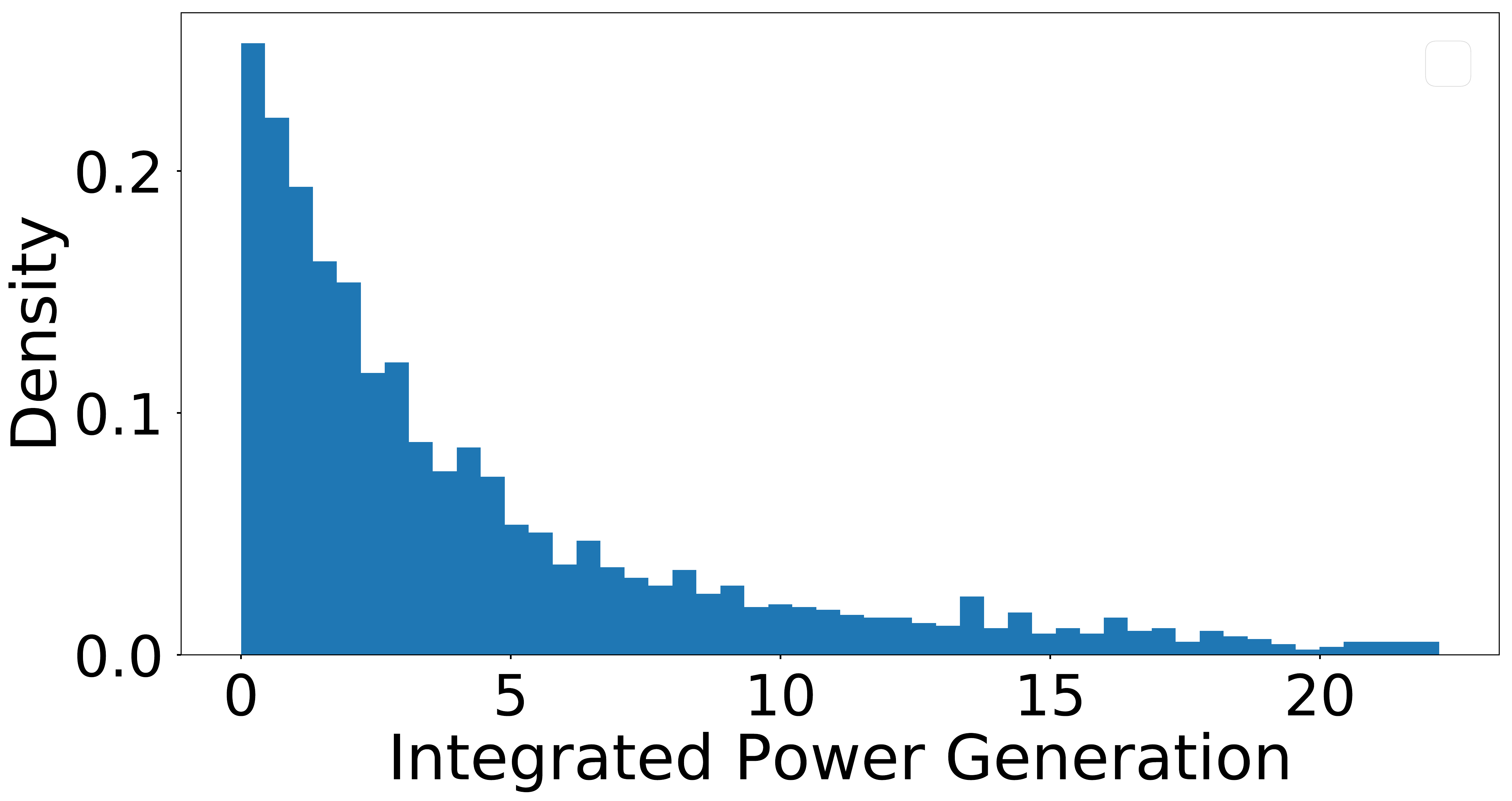} }}%
    \hfill
    \caption{Integrated power generation from historical data of the EuropeWindFarm2015 dataset to asses the amount of stress within the considered horizon. A maximum value of $24$ is possible as maximum stress level.}%
    \label{integration_wind}
\end{figure}

\begin{figure}[!tb]%
    \centering
    \subfloat[Integrated power generation from historical data.]{{\includegraphics[width=0.48\textwidth]{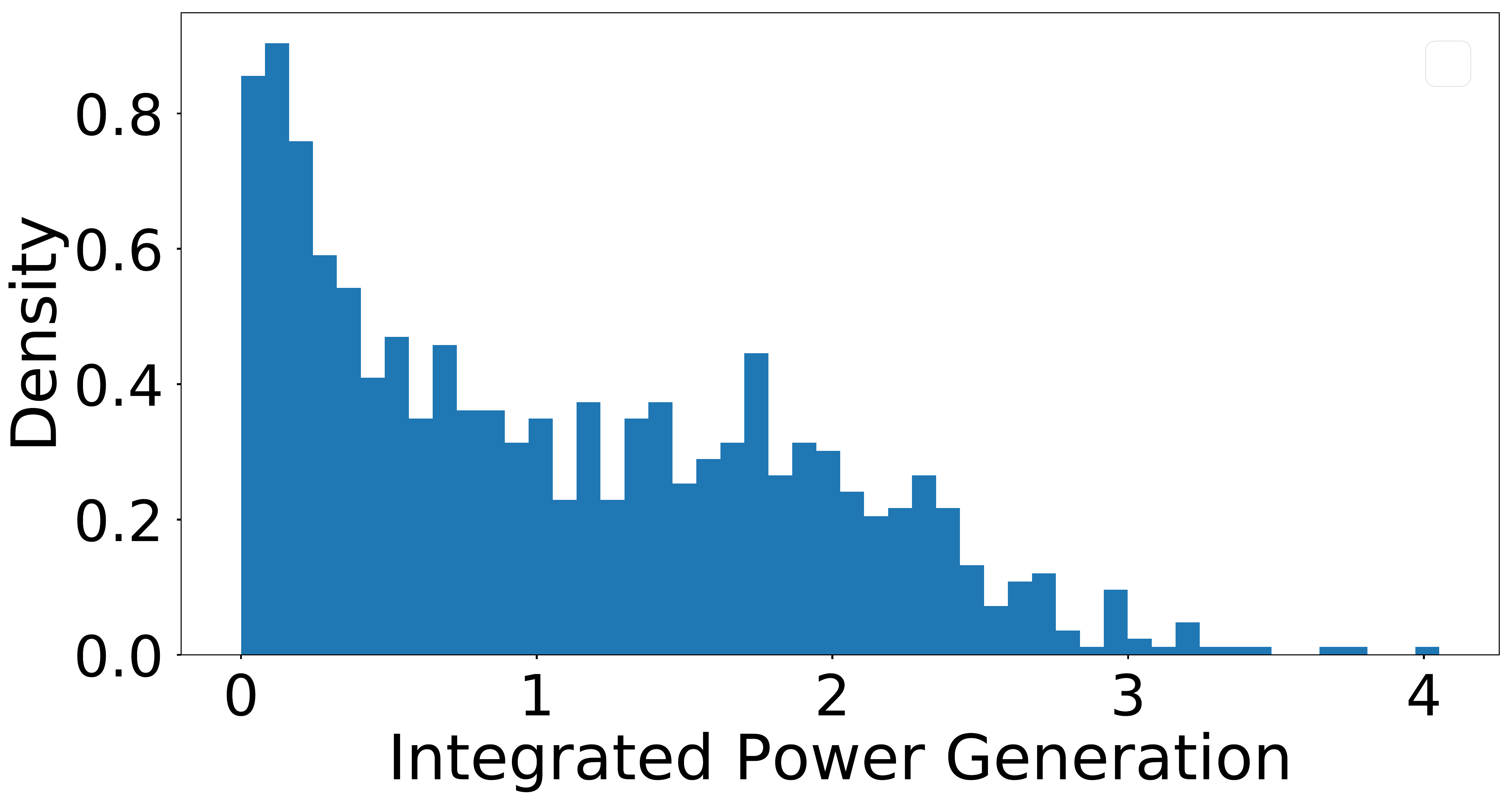} }}%
    \hfill
    \subfloat[Integrated power generation from generated DC-WGAN samples.]{{\includegraphics[width=0.48\textwidth]{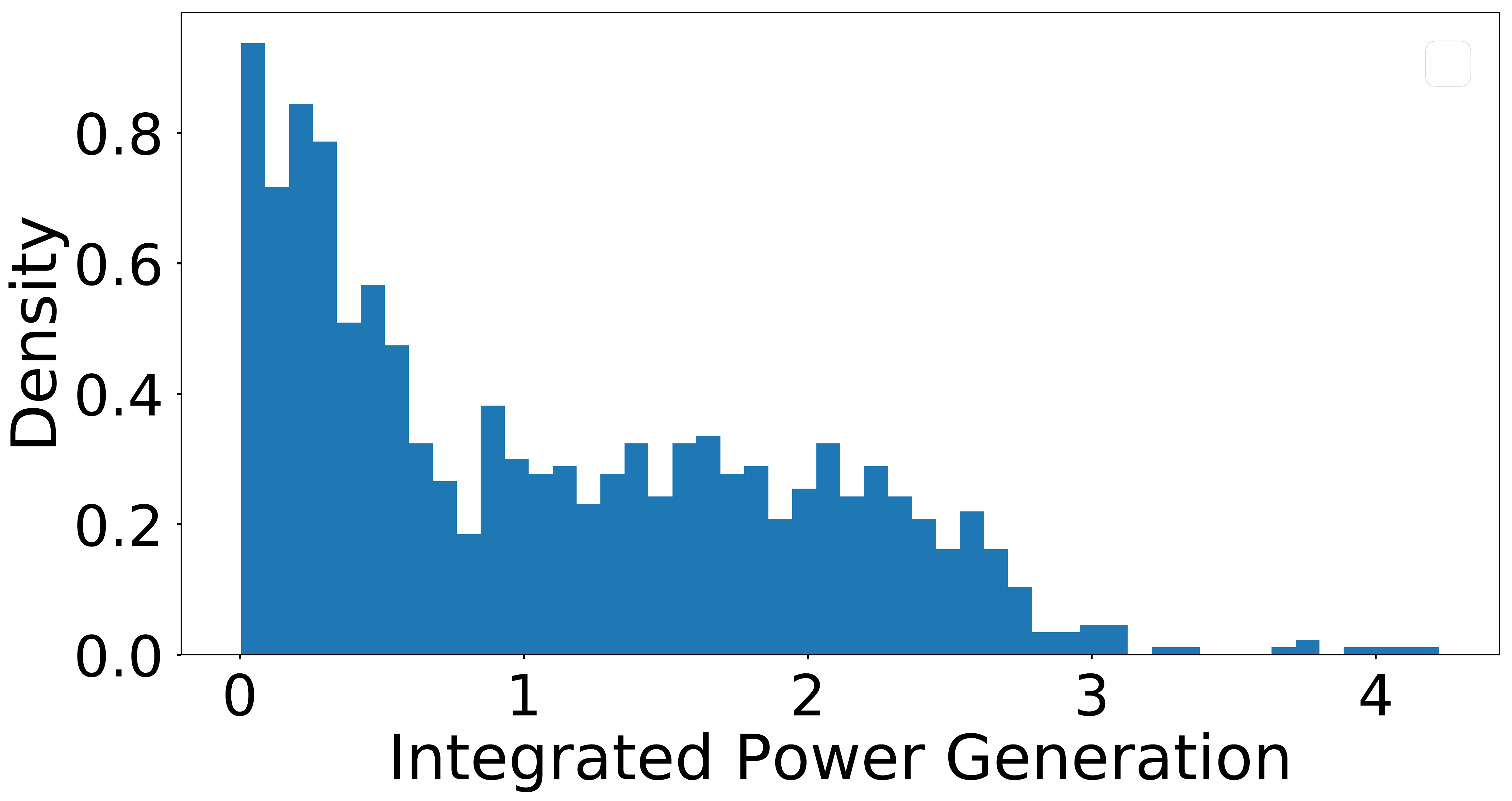} }}%
    \hfill
    \caption{Integrated power generation from historical data of the GermanSolarFarm2015 dataset to asses the amount of stress within the considered horizon. A maximum value of about $4$ is possible as maximum stress level.}%
    \label{integration_pv}
\end{figure}

Figure~\ref{integration_wind} and \ref{integration_pv} provide examples of this analysis for the EuropeWind\-Farm\-2015 and the GermanSolarFarm\-2015 datasets summarized by histograms. 
Both results show that the generated stress level is similar to the one of the historical data. 
However, due to the small amount of historical data with high-stress situations, the generated samples contain mostly values below a value of $10$ for wind and about $2.5$ for solar.
The latter results suggest that there are only a small amount of days with intense solar radiation throughout the whole day.
The former relates to the fact that wind farms typically are ramped down when working at a maximum level over a long period.

\subsection{Study on Location-Specific Distribution Generation}
The following study reveals how location-specific influences and their location-specific wind conditions (that affect the power distribution) are modeled by the GANs and GCs when trained simultaneously with different terrains. 
The evaluation is similar to the previous section but omits the analysis of spatial and temporal relationships as results are identical to the previous study.

\begin{figure}%
    \centering
    \subfloat[PDFs of complete dataset.\label{real_all_pdf}]{{\includegraphics[width=0.99\textwidth]{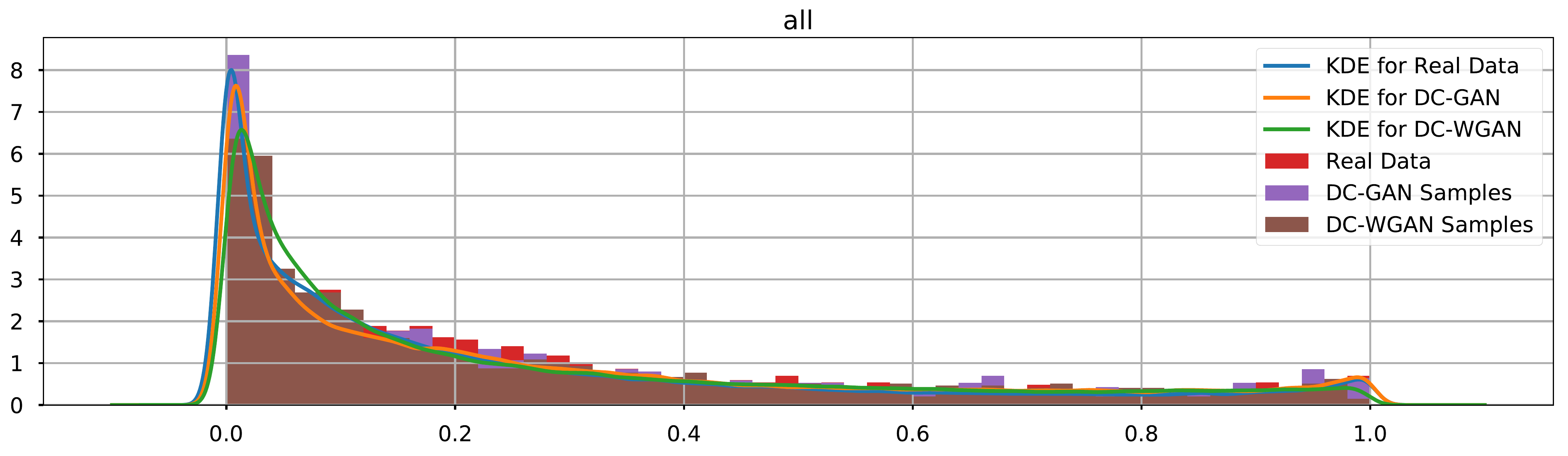} }}%
    \\
    \subfloat[PDFs of flatland terrain.]{{\includegraphics[width=0.99\textwidth]{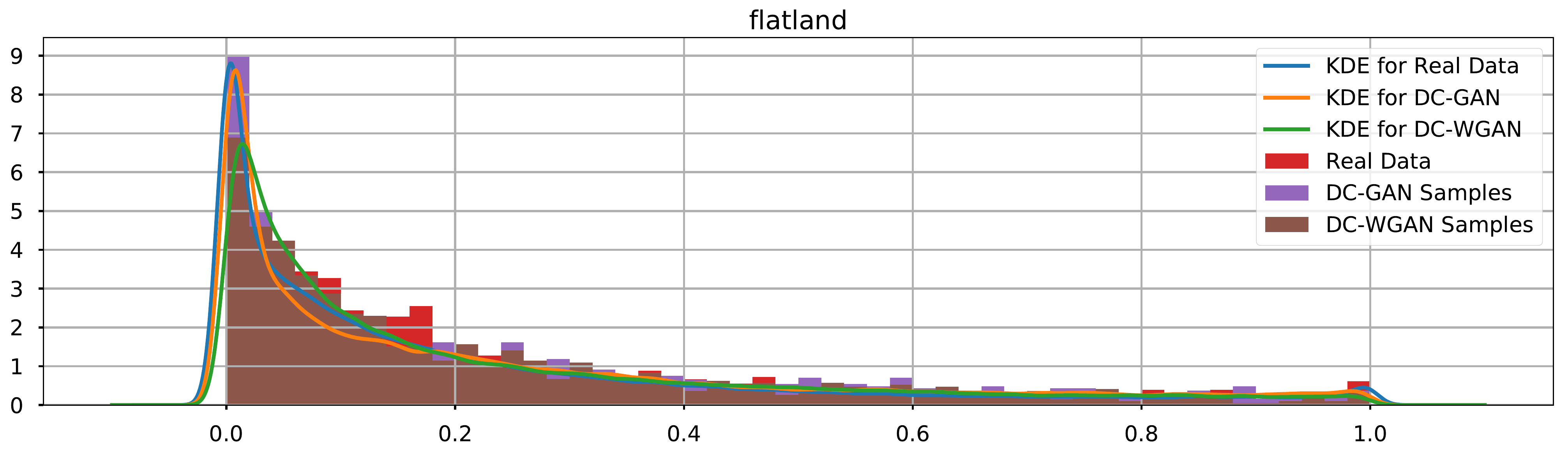} }}%
    \\
    \subfloat[PDFs of forest terrain.]{{\includegraphics[width=0.99\textwidth]{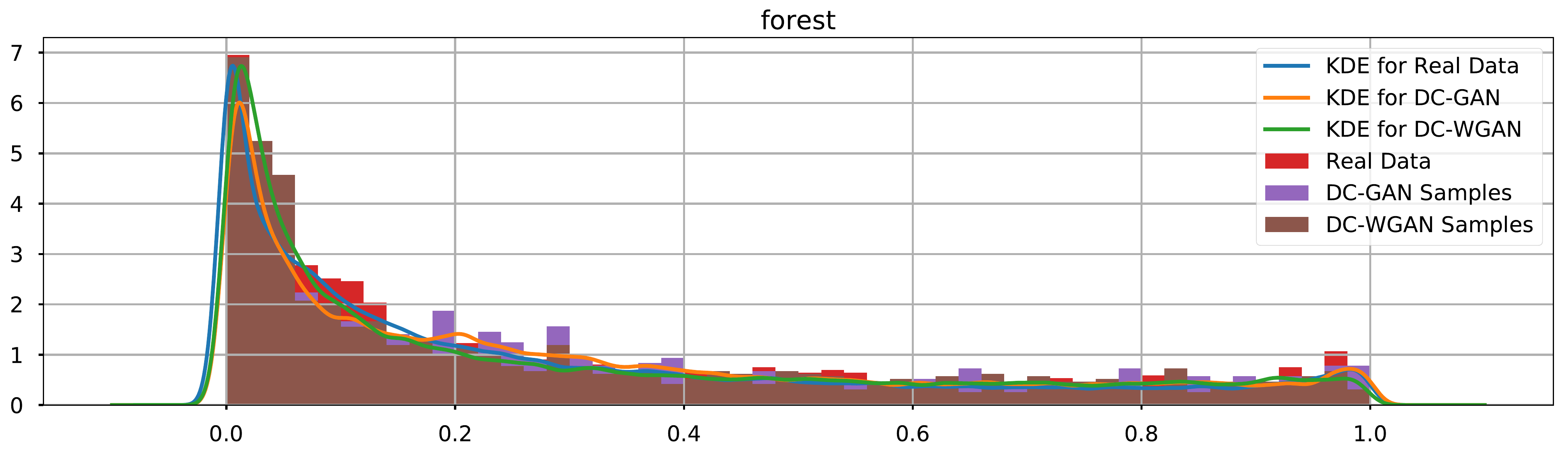} }}%
    \\
    \subfloat[PDFs of offshore terrain.]{{\includegraphics[width=0.99\textwidth]{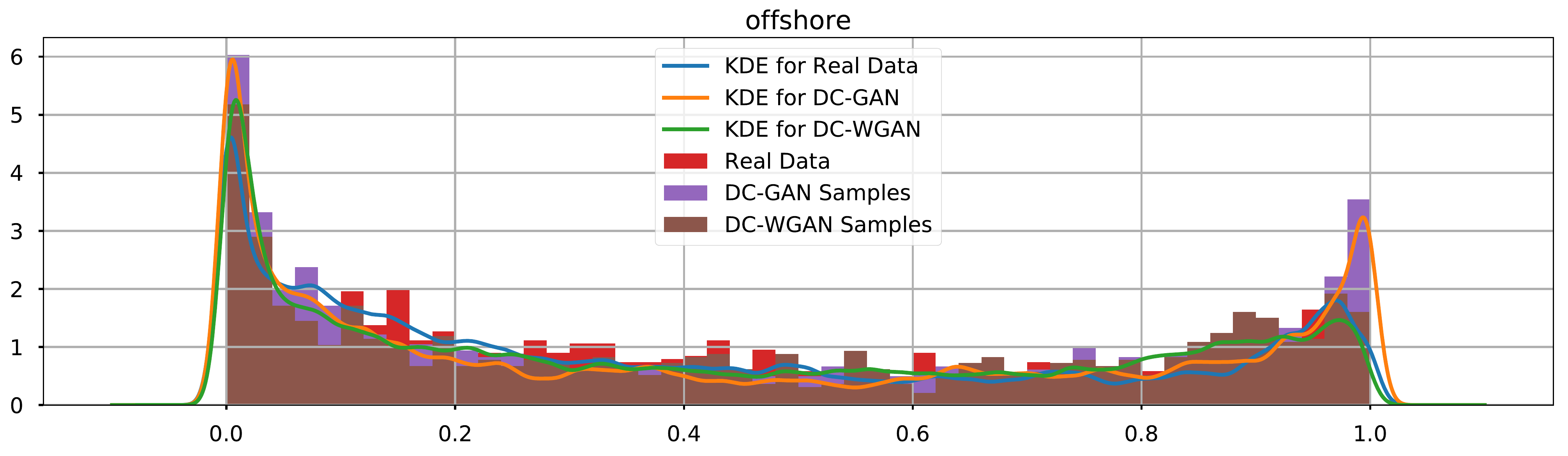} }}%
    \\
    \caption{The different figures compare PDFs approximated using a KDE for the real and generated data of the GermanWindFarm2017 dataset for each terrain and all farms.\label{locations}}
\end{figure}

In Table~\ref{Distributions}, we compare the distributions of the generated samples for all models.
We calculate the KLDs between the distributions of all real samples and all generated samples. By grouping farms by their terrain, we estimate the location-specific KLD. 
The results show that both models create a distribution similar to the historical data (also compare Figure~\ref{locations}). 
In cases of flatland, the DC-GAN has a smaller KLD, for the forest terrain the values are equal, and for offshore farms, the DC-WGAN has a smaller KLD.
Again, the GC achieves smaller KLD values compared to the DC-GAN but larger amounts compared to the DC-WGAN.

\begin{table}
    \centering
        \begin{tabular}{|l|c|c|c|c|c|c|}
        \hline
        Location              & KLD GC            &KLD DC-GAN        &KLD DC-WGAN & \#Farms     \\
        \hline
        \hline
        Flatland              & 0.143            &0.194            &    0.037       &32                    \\
        Forest                & 0.085             &0.266            &    0.018       &10                    \\
        Offshore              & 0.148              &0.304           &    0.046       &4                    \\
         \hline
        \end{tabular}
        \vspace{1em}
    \caption{The table shows the similarity measured by the KLD for terrain specifics power distributions from the GermanWindFarm2017 dataset. In particular outcomes of the offshore and forest terrain are relevant as those have a limited amount of data.}
\label{Distributions}
\end{table}
%
In Figure~\ref{locations}, it can be seen that for each terrain and GAN the PDFs are similar to the historical data. 
For simplicity, we omit the presentation of the GC as we are interested in evaluating GANs for renewable power generation.

The following analysis refers to values from historical data. Both GANs create similar values. However, results of the DC-WGAN are closer to the test dataset.
For all terrains, a higher density occurs in the low yield range. 
The density decreases further for increased yields but rises again slightly in the range of maximum power. 
Wind farms on flatland have a mean at $0.201$.
Wind farms near forests have an increased mean at $0.263$ in the historical data, resulting in a higher total yield.
For offshore wind farms, the average power generation is $0.381$, with a higher share in the range of maximum power. 
Similar differences between the terrains are present in the variance and skewness. 
Flatland has the most remarkable skewness value, forest the second largest, and offshore the smallest one. 
The order is the opposite in the magnitude of variance values.

Results of the study confirm that the GANs are capable of modeling the terrain specific power distributions due to site-specific wind conditions even for a limited amount of data as for offshore and the forest terrains. 
In particular, the GANs create similar PDFs specific to those terrains. 

\subsection{Discussion}

Interestingly, even when training GANs on, e.g., only $16$ solar farms and a widely varying amount of capacities, the results of the KLD show that the generated power distribution is similar to historical data. 
The representative histograms in Figure~\ref{locations} confirm those similarities.

Also, in the study of terrain specific distributions, both GANs learn the individualities of each terrain. 
Statistical values such as mean, variance, and skewness are closer to historical in samples from the DC-WGAN. 
Impressively in created data of the offshore territory, the large density, in the range of maximum power is also captured by both GANs. 
This effect might be due to simultaneous learning (similar to a multi-task approach) of the different farms allowing to capture small individualities for each farm and their site-specific conditions.

A critical remark of the analysis needs to be done concerning seasonal effects, which is challenging to consider due to the limited amount of data.
Another problem is related to the number of high-stress situations. 
Due to the limited occurrences in the historical data, the chance of creating those by the GANs are also low.
However, they can be created through repeated sampling and rejecting those below a certain threshold of the integrated power value.

Overall, the DC-WGAN is superior in modeling the spatial and temporal relations as well as power distributions even when faced with limited data compared to previous studies. 

\section{Conclusion and Future Work}
\label{conclusion}
In this article, we compared the binary-cross-entropy (BCE) loss and the Wasserstein distance to the training of GANs on four different data sets. 
Results show the superior quality of the Wasserstein distance over the BCE loss and a GC as the baseline to generate the power distribution when taking spatial and relationship and the KLD into account.
The publicly available source code, the datasets, and the results provide a basis for comparison when utilizing GANs in the scope of renewable scenario generation.
Ultimately, we confirmed that GANs are capable to model different power distributions, including external influences such as terrains even when faced with a limited amount of data compared to previous studies. 

A future goal is to utilize GANs to impute missing values or create samples for unknown farms by creating a GAN conditioned on weather events and previous power values. 
The latter case also allows using the samples in the field of transfer learning.
\\
\newline
\small
\textbf{Acknowledgment:}
This work was supported within the project Prophesy (0324104A) funded by BMWi (Deusches Bundesministerium für Wirtschaft und Energie / German Federal Ministry for Economic Affairs and Energy).\\
Additionally, special thanks to Maarten Bieshaar for excellent discussions about Gaussian copulas.
\vspace{-0.5em}
\bibliographystyle{unsrt}
\bibliography{references.bib}

\end{document}